# Automatic image annotation base on Naïve Bayes and Decision Tree classifiers using MPEG-7

Jafar Majidpour
Dept. Computer Science, University of Raparin
Rania, Iraq
Jafar.majidpoor@uor.edu.krd

Samer Kais Jameel
Dept. Computer Science, University of Raparin
Rania, Iraq
Samer.kais@uor.edu.krd

*Abstract*—recently It has become essential to search for and retrieve high-resolution and efficient images easily due to swift development of digital images, many present annotation algorithms facing a big challenge which is the variance for represent the image where high level represent image semantic and low level illustrate the features, this issue is known as "semantic gab". This work has been used MPEG-7 standard to extract the features from the images, where the color feature was extracted by using Scalable Color Descriptor (SCD) and Color Layout Descriptor (CLD), whereas the texture feature was extracted by employing Edge Histogram Descriptor (EHD), the CLD produced high dimensionality feature vector therefor it is reduced by Principal Component Analysis (PCA). The features that have extracted by these three descriptors could be passing to the classifiers (Naïve Bayes and Decision Tree) for training. Finally, they annotated the query image. In this study TUDarmstadt image bank had been used. The results of tests and comparative performance evaluation indicated better precision and executing time of Naïve Bayes classification in comparison with Decision Tree classification.

*Keywords-component; EHD, SCD, CLD, Decision Tree, Naïve Bayes*

## I. INTRODUCTION

Lots of images are produced from many references such as architectural and engineering, advertising, and medical images [1]; consequently the challenge is how efficiently use this huge amount of data [2], therefore, necessary to have a technique able to retrieve those images and solve the issue, "Image annotation or automated image annotation" is the importance way to resolve the issue especially this technique recover the content and semantic concept for images [3]. In the annotation process, it is important and necessary to retrieve the image content in high accurately and swiftly, so it is necessary to use Content-Based Image Retrieval (CBIR) system for this purpose [4], which illustrated in figure 1. Most of the techniques that are used to retrieve the images had failed to resolve semantic gap issue due to it adopt on color, texture, shape, spatial layout that is considered as low level features [5]. semantic gap problem is defined by Some researchers as "lack of coincidence between the information that one can extract from the visual data and the interpretation that the same data have for a user in a given situation" [6], as shown in figure 2.

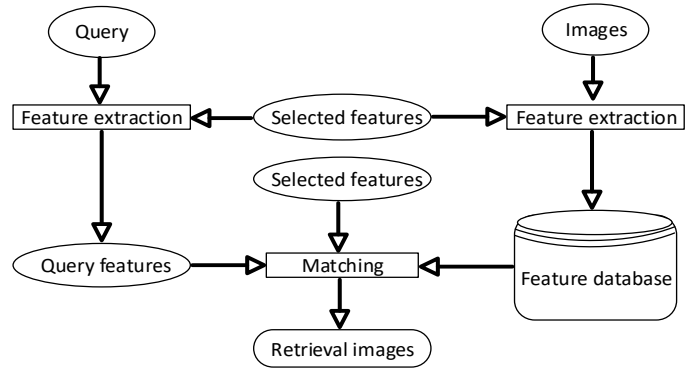

Figure 1. CBIR system

Some techniques have capabilities that enable them to overcome this trouble for instance machine learning, object ontology, and semantic template[7], That is why the system presented in this paper is proposed, which is part of the machine leaning technique, to cover vulnerabilities found in previous methods to some extent.

Figure 2. Semantic Gap illustration

## I. RELATED WORK

*A.* Monolisa, et al had used feature extraction (SIFT algorithm) followed by clustering algorithm in order to image annotation mechanism. [9].

*B.* Tianxia, et al used clustering algorithm attention for annotation, two sets of candidate to personalizing the automatic image annotation [10].

*C.* Yiren Wang, et al; they have compared 6 kernels, and analyzed methods by conducting intensive experiments to examine performance of feature mapping [11].

*D. Haijiao Xu, et al,* in this work mixture links has employed between Google concept and local concept, and semantic links between single-concepts and multi-concepts [12].

*E. Chenyi Lei; et al,* they propose a framework for image annotation through learning rank, and the use real dataset for extract feature that is by employing the records of social diffusion, the result shows that the proposed system is better than other systems [13].

*F.* Lixing Jiang, et al, used Decision Tree-based Naïve Bayes for automatic annotation, by take advantages of both Decision Tree and Naïve Bayes [14].

## II. THE PROPOSED APPROACH

Increasing the precision of image annotation is one of the necessities of minimizing the above-mentioned problem. Since the semi-automated method requires a great deal of time and high cost, for this reason, it has become a pressing need to propose an automated system that helps reduce cost and increase the accuracy of the results in addition to reducing the execution time. Some tools also has been planned for improve the precision of extracting features from images.

The proposed criteria for improving annotation encompass increasing precision, reducing the dimensions of extracted features, system performance, training speed, and reducing executing time.

To reach high level semantic perception, machine learning techniques were used e.g. Naïve Bayes and Decision Tree that contributed to making the system automatic and reduce the semantic gap.

To evaluate the efficiency of the system there must be a standard of measurement, which is accuracy. The system is training by using training set images to capture the features that enable it to retrieve the content of images, and then using other dataset called test set and calculate the accuracy that is ratio of the images that are correctly annotated over the total image annotated as shown in equation (1).

$$\text{Accuracy} = \frac{\text{Correctly annotated images}}{\text{Total images annotated}} \quad (1)$$

Another evaluation criteria is executing time which is the time required to perform an automated annotation on the query images.

*A. Automated Image Annotation system*

Before starting to implement the automated image annotation system, the images were placed in folders (classes) having relevant subjects. Each folder represents a class of the same subject.

EHD used to extract the texture features from the image and produced a feature vector with 80 elements, SCD employs to produce color feature vectors consist of 256 elements, and CLD has resulted an array of three rows and 64 columns elements, this array passed to (PCA) to convert into vector with 64 elements.

In the process of extracting features, the vectors are produced, which are considered input to the classifiers (naïve Bayes and Decision tree) in order to begin the system training process.

Now that the training process has been completed using the training dataset, the system is ready to receive the test images to be classified into a specific category. Finally, the automatic annotation process has been completed, which is shown in the figure 3 below.

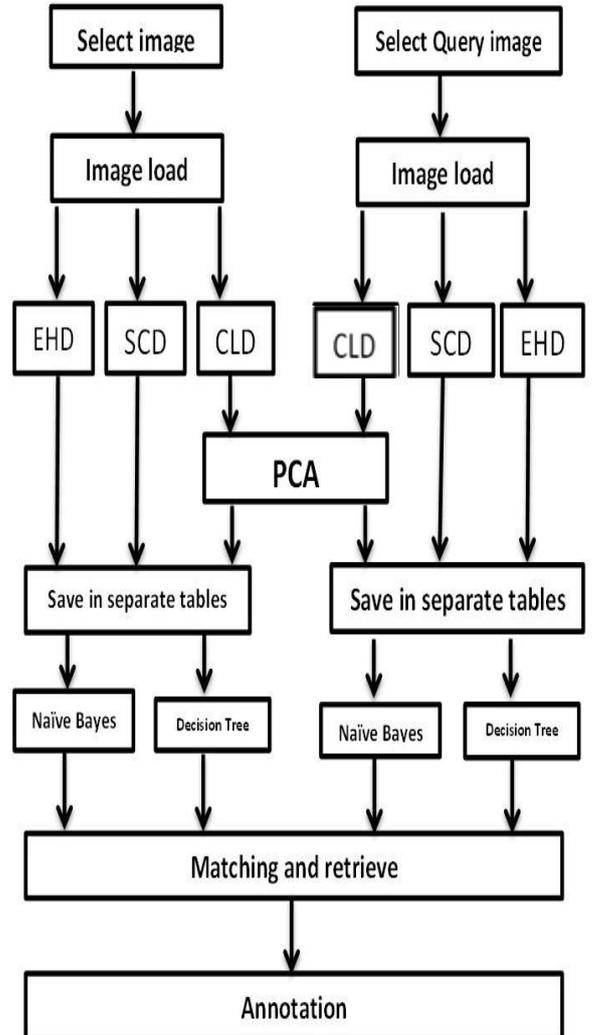

Figure 3: Automatic image annotation proposed

## B. Feature Extraction

Images contain characteristics that can distinguish this image and thus give the ability of the classifier to belonging the image into its class, based on is its content. These characteristics are present in the image but in a scattered and repetitive manner. This is the role of extracting information from these data, and saved it in the form of vectors in order to contribute to reduce the time of image retrieval and facilitate the process of classification [15]. The more accurate the process of extracting information, the more accurate the classification process will be [16]. The training set will change depending on the feature extraction algorithm; the degree of change is represent the robustness of algorithm [17].

### 1) Characteristic Extraction of EHD

Extraction the edge is an aspect task as it provides us with information on the behavior of the edges in content of image. The edge histogram descriptor analyzes the images and provides information about the edges in the images in form of vector. the edges are described in five different directions [18]

In the beginning, the image is divided into eight sub-images, each sub-image is processed individually, the descriptor creates a histogram that includes five pinpoint, which represents the value for the five different edges directions for all eight sub images, the first direction is the vertical direction, second is horizontal, third is diagonal with 45 degree of angle, the fourth is diagonal with 135 angle degree, and the last one is non-directional [19]. To construct the histogram the sub-image in turn divided into blocks, and each block is partition into four sub-blocks and labeled from 0 to 3 as shown in figure 4, the edge is extracted by applying a digital filter, pin histogram value normalize by the total number of blocks [18].

The value of horizontal edge calculated by $S_0 = \left| \sum_{k=0}^{3} ak(i,j) f_h(K) \right|$, for vertical edge $S_{90^0} = \left| \sum_{k=0}^{3} ak(i,j) f_v(K) \right|$, diagonal edge with 135 angles $S_{135^0} = \left| \sum_{k=0}^{3} ak(i,j) f_{135}(K) \right|$, diagonal with 45 degree of angle $S_{45^0} = \left| \sum_{k=0}^{3} ak(i,j) f_{45}(K) \right|$, and non-directional $S_{nd} = \left| \sum_{k=0}^{3} ak(i,j) f_{nd}(K) \right|$

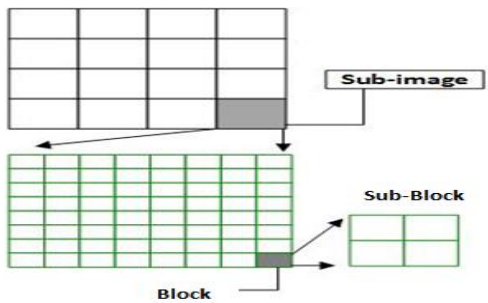

Figure 4. image and Sub-image

### 2) Scalable Color Descriptor (SCD)

Used HSV color space with fix quantization to generate histogram with 256 pins to achieves full interoperability between different resolutions of the color representation [20]. SCD stated in frequency domain by use Haar transform (simplest energy compression process [4]) in histogram, SCD stated from low frequency to high frequency [4]. Figure 5 indicate to procedure for extracting color feature.

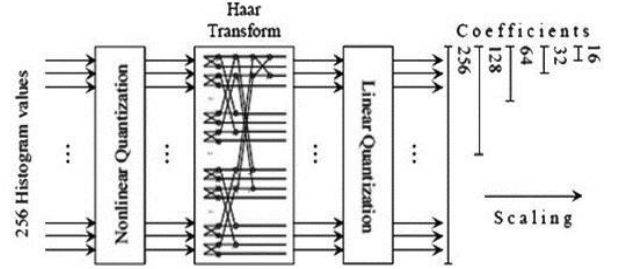

Figure 5. Diagram of the feature extraction process from SCD [4]

### 3) Color layout descriptor CLD

This type of descriptor defines the color feature in the images [21], Before the start procedure of extraction, image converted to YCbCr color space, and then employ the discrete cosine transform (DCT) after dividing the image into 64 sub-images, the resulting from DCT is a set of coefficients saved in a vector containing values describing the colors of the image [22, 23]. Figure 6 illustrations the algorithm of CLD.

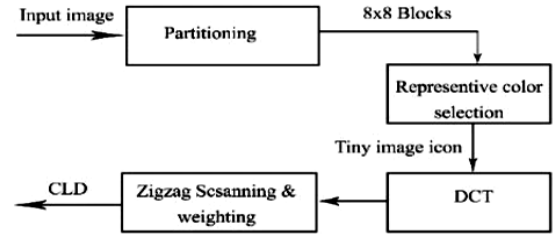

Figure 6. block diagram of the CLD extraction [4]

## C. Machine learning

It is technique that use statistical model to analyze data and support the system to able to training the data in order to increase the performance of decision making [24].

### 1) Naïve Bayes classifier

In machine learning especially researches that depend on classification, Naïve Bayes method is simplest and more effective, sometime it is optimal in some situation [24].in addition, it based on statistical models founded on Bayesian theorem [25]. Uses certain attributes that do not depend on each other, and classified under predefined condition [26], For the training process, this classifier does not need large training dataset, as it do the classification with high accurate even that the data is not large, this is one of the characteristics of this classifier, this classifier is used in AIA (Automatic Image Annotation). Conditional model can signify by.

$$P(Y | X_1,...., X_n) \qquad (2)$$

Naïve Bayes calculating is

$$\text{posterior} = \frac{\text{prior} * \text{likelihood}}{\text{Evidence}} \quad (3)$$

That is:

$$P(y | x_1...x_n) = \frac{P(y) * P(x_1...x_n | y)}{P(x_1...x_n)} \quad (4)$$

*2) Decision Tree classifier*

Is one of classification techniques [27], characterized by its ability to classification easily, and its ability to deal with recurring features [28]. This algorithm starts with root configuration and then creates the rest of the branches of the tree. The output is a tree, each part of which represents a decision scenario to classification process [29]. This tree used in the process of classification, where the image is processed in succession sequence from the root to the leaves through the branches in order to make a decision and classify query image into one of the categories [30]. Figure 7 illustrate an example of Decision Tree.

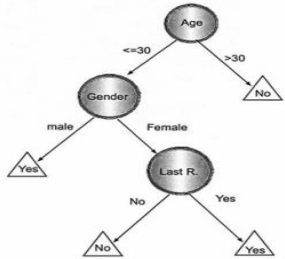

Figure 7. example of Decision Tree [31]

*3) C4.5 Algorithm*

This algorithm is most common use for classification purpose [32]. Here each node in the tree tested by testing attributes of node and the attribute that contains the largest gain is chosen as the attribute for that node [33]. The gain is calculated by entropy. One of the dataset is $A_i$ that is belong to dataset D which is content set of classes $C_1$, $C_{12}$, ……, $C_m$. therefore the gain calculated as [35].

$$\text{Gain}(D, A_i) = \text{Entropy}(D) - \sum_{i \in \text{Value}(A_i)} \frac{|D_i|}{D} \text{Entropy}(D_i) \quad (5)$$

Where,

$$\text{Entropy}(D) = -\sum_{i=1}^{m} \frac{\text{freq}(C_i, D)}{|D|} \log_2 \frac{\text{freq}(C_i, D)}{|D|} \quad (6)$$

Freq ($C_i$,D) : the number of examples in the dataset D belong to class $C_i$ and $D_i$ divided by attribute $A_i$s value $a_i$.

Then SplitInfo ($A_i$) is define as the information content of the attribute Ai itself [35].

$$\text{SplitInfo}(D, A_i) = -\sum_{i \in \text{Value}(A_i)} \frac{|D_i|}{D} \log_2 \frac{|D_i|}{D} \quad (7)$$

The gain ratio is the information gain calibrated by splitInfo:

$$\text{Gain Ratio}(D, A_i) = \frac{\text{Gain}(D, A_i)}{\text{SplitInfo}(A_i)} \quad (8)$$

### III. SIMULATION

This paper was simulated using C#. It included two phases, which were feature extraction phase and classifier-training phase. Moreover, TUDarmstadt image bank has used, which are set of images consist of 325 images with three classes, first one is related with motorcycles, second is about cars images, and last one is consist of cows images. Tests were conducted in 10 phases, in each of which 295 images i.e. 102, 90, and 103 for motorcycle, cars, and cows respectively, these images have given to Naïve Bayes and Decision Tree classifier for training. Features of images has been extracted and saved in separate tables via employing EHD, CLD, and SCD algorithms. these algorithms did the automated annotation and returned the annotation results including the relevant class name accompanied by a percentage error. Table 1 and figure 8 indicates the annotation results using Naïve Bayes and Decision Tree classifiers with feature extraction algorithms separately, and table 2 with figure 9 indicates results of execution time for each classifier with three descriptor separate states.

Table 1. results of precision for each classifier with three descriptor separate states (%)

| Descriptor / Classifier | Naïve Bayes | Decision Tree |
|---|---|---|
| EHD | 94.4 | 83.7 |
| CLD | 70.8 | 64.6 |
| SCD | - | 81.5 |

Table 2. results of execution time for each classifier with three descriptor separate states (seconds)

| Descriptor / Classifier | Naïve Bayes | Decision Tree |
|---|---|---|
| EHD | 0.326 | 13.383 |
| CLD | 0.280 | 10.854 |
| CSD | 2.372 | 27.626 |

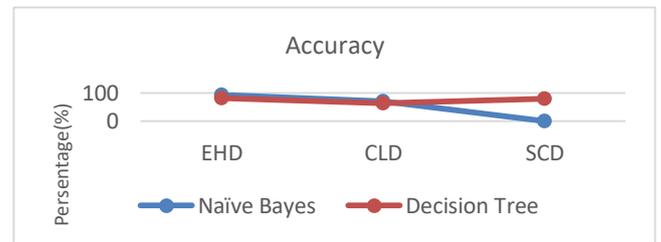

Figure 8. The curve precision in Naïve Bayes and Decision Tree classifiers

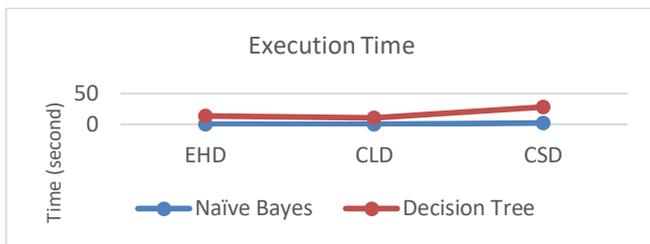

Figure 9. The curve execution time in Naïve Bayes and Decision Tree

## IV. COMPARISON AND EVALUATION

### A. Precision

*1)* Observely, that results of Naïve Bayes annotation were better than those of Decision Tree, because Has the ability to use all the features extracted in the training phase and compare them with the features that were extracted in the testing phase. However, they were compared with less than a half of trained features in Decision Tree method. Therefore, the precision of Naïve Bayes annotation was higher.

*2)* Features of images which were extracted with EHD algorithm, were annotated wrongly because some images were taken from a farther distance. It means that the targeted object was placed in a farther distance so that Naïve Bayes and Decision Tree classifiers annotated such images wrongly.

*3)* EHD algorithm would use the image texture for feature extraction; however, CLD and SCD algorithms would use the image color. In some images from different classes, the similarity of subject colors caused CLD and SCD algorithms to have weaker performances in comparison with EHD algorithm

*4)* CLD algorithm used the representative color, for each part, the average color is calculated for each pixel for feature selection. some cases, the background color was selected for the representative color. It reduced the precision of classifiers in annotation; for instance, it discard the object color, which it should consider for compare, and look to the background color for comparizm purpose.

*5)* SCD algorithm collect the characteristics of the image it are like a 256-element array of real numbers. The Naïve Bayes classification used a normal distribution.. The Naïve Bayes classification used a normal distribution. Due to the fact that normal distribution selected continuous numbers, and the features extracted from SCD algorithm were discrete, the annotation results were wrong. It means that it would annotate only the first class for all images (all subject).

### B. Executing time

The executing time of tested image in Naïve Bayes classifier was lesss in comparison with Decision Tree classifier because the calculation process of information gain in C4.5 algorithm involved the compute of log function, and the time consuming of the log function is a problem on total computation time.

## V. CONCLUSION

The first problem is the problem of different representation of attributes between the high and low level, which is called semantic gap, the second problem is in the training phase, that the words do not match with the image regions, to bridging the semantic gap can be used machine learning techniques. In this paper, texture and color features are used to support image retrieval with high accuracy,reduce annotation execution time, growth system competence, Solve the problem of increasing dimensions via reducing them, and training the data swiftly.